\RequirePackage{fix-cm}
\documentclass[smallcondensed]{svjour3}     % onecolumn (ditto)
\smartqed
\usepackage{graphicx}
\usepackage[square]{natbib}
\usepackage{algorithm}
\usepackage{algorithmic}
\usepackage{amsmath}
\usepackage{amssymb}
\journalname{Machine Learning}
\begin{document}

\title{Inverse Reinforcement Learning from Summary Data}

\titlerunning{IRL from Summary Data}

\author{Antti Kangasr\"a\"asi\"o \and
        Samuel Kaski
}

\institute{Antti Kangasr\"a\"asi\"o \at
              Aalto University
              Department of Computer Science
              \email{antti.kangasraasio@iki.fi}
           \and
           Samuel Kaski \at
              Aalto University
              Department of Computer Science
              \email{samuel.kaski@aalto.fi}
}

\date{Received: date / Accepted: date}

\maketitle

\begin{abstract}
Inverse reinforcement learning (IRL) aims to explain observed strategic behavior by fitting reinforcement learning models to behavioral data.
However, traditional IRL methods are only applicable when the observations are in the form of state-action paths.
This assumption may not hold in many real-world modeling settings, where only partial or summarized observations are available.
In general, we may assume that there is a summarizing function $\sigma$, which acts as a filter between us and the true state-action paths that constitute the demonstration.
Some initial approaches to extending IRL to such situations have been presented, but with very specific assumptions about the structure of $\sigma$, such as that only certain state observations are missing.
This paper instead focuses on the most general case of the problem, where no assumptions are made about the summarizing function, except that it can be evaluated.
We demonstrate that inference is still possible.
The paper presents exact and approximate inference algorithms that allow full posterior inference, which is particularly important for assessing parameter uncertainty in this challenging inference situation.
Empirical scalability is demonstrated to reasonably sized problems, and practical applicability is demonstrated by estimating the posterior for a cognitive science RL model based on an observed user's task completion time only.
\end{abstract}

%%%%%%%%%%%%%%%%%%%%%%%%%%%%%%%%%%%%%%%%%%%%%%%%%%%%%%%%%%%%%%%%%%%%%%%%%%%%%%%%%
\section{Introduction}\label{sec:intro}

Inverse reinforcement learning (IRL) has generally been formulated \citep{russell1998learning,ng2000algorithms}
as:\\
\textbf{Given} (1) a Markov decision-process (MDP) with reward-function $R(s; \theta)$, where the $\theta$ are unknown parameters;
(2) a set of state-action paths $\Xi = \{\xi_1, \dots, \xi_N\}$ demonstrating optimal behavior given the true $\theta^*$, where $\xi_i = (s^i_0, a^i_1, \dots, a^i_{T_i-1}, s^i_{T_i})$;
and optionally (3) a prior $P(\theta)$.\\
\textbf{Determine} a point estimate $\hat{\theta}$ or the posterior $P(\theta|\Xi)$.

%Methods for solving the IRL problem have been used for parameter inference in multiple real-world modeling situations where complex behavior has been observed in the form of state-action paths.
%Examples include
%driver route modeling \citep{ziebart2008maximum},
%helicopter acrobatics \citep{abbeel2010autonomous},
%learning to perform motor tasks \citep{boularias2011relative},
%dialogue systems \citep{chandramohan2011user},
%pedestrian activity prediction \citep{ziebart2009planning,kitani2012activity}, and
%commuting routines \citep{banovic2016modeling}.

IRL problems arise when it is of interest to infer the goals or predict future behavior of intelligent agents based on observations of the agent's past behavior.
Overall, there are many situations where humans behave in a complex and adaptive manner, which might not be explainable by a simpler model.
Examples include
driver route modeling \citep{ziebart2008maximum},
helicopter acrobatics \citep{abbeel2010autonomous},
learning to perform motor tasks \citep{boularias2011relative},
dialogue systems \citep{chandramohan2011user},
pedestrian activity prediction \citep{ziebart2009planning, kitani2012activity}, and
commuting routines \citep{banovic2016modeling}.

Humans are, in general, able to understand and predict the behavior of other humans in familiar settings, even from rather limited observation data.
Developing a similar ability in autonomous agents could thus, for example, enable them to interact more naturally with humans in rich every-day situations.
However, a limitation with the traditional problem formulation is the assumption that full paths containing both actions and states have been observed.
In many real-world situations such fine-grained observations may not be available for multiple reasons.
For example, it may be too costly to set up sensors that could gather the fine-grained observations, or it may even be impossible to change the measurement devices if they are owned by a third party.
Also, even if accurate sensors are used, various environmental factors may cause unavoidable occlusion, censoring or distortion to the measurements.
Furthermore, existing datasets are unlikely to contain full path data, if the data have not been collected with IRL in mind.
We elaborate on these motivations later.
%Reasons include cost of measuring and storing full paths, environmental restrictions such as occlusion, privacy considerations, and adversarial actions.
%the number of cases where only partial observations are available is probably much larger than those with accurate path data.

There have been a few initial approaches for addressing this issue.
The earliest was to assume that instead of the actual paths, we might just observe the expected sum of state feature values the agent encounters during the demonstrated paths, known as feature expectations \citep{abbeel2004apprenticeship}.
Later approaches have relaxed the assumption on the state observations from accurate to probabilistic: instead of observing the states, they assume a probability distribution $P(s_t)$ over the state-space is given for each timestep \citep{kitani2012activity}.
However, the existing methods are not applicable in more general situations, where the external observer has partial observability at the path level.

\textbf{Summary of contributions:}
This paper formulates the IRL from summary data (IRL-SD) problem, which extends the IRL problem to situations where the full paths are not directly available.
We assume a summarizing function $\sigma$ acts as a filter between the external observer and the true paths.
We demonstrate that even in the most general case with no prior assumptions about the summarizing function, inference is still possible for this problem class, thus significantly extending the scope of problems where IRL can be performed.
We derive the exact likelihood for this problem and two approximations that are significantly faster to evaluate.
The first approximation is a Monte-Carlo estimate and the second uses an approximate Bayesian computation (ABC) approach.
We demonstrate that both of these approximations are feasible for MDPs for which optimal policies can be estimated in a reasonable time.
Using a grid world toy example, we demonstrate that both the exact and approximate methods are able to recover the parameters of the reward function with good accuracy, and that the approximate methods scale significantly better.
Using a recent RL model from the cognitive science literature, we demonstrate that a sensible approximate posterior can be inferred based only on the task completion times collected from user experiments.

The methods have additional interesting properties.
First, they do not differentiate between different types of MDP parameters, which allows inference to be easily extended to any interesting parameters of the generative process besides the traditional reward function.
Second, they also allow non-linear reward functions to be used, which is not the case with many existing methods.
Third, the approximate methods can also be used in situations where the transition function is not known, as long as we can generate draws $s_{t+1} \sim P(s_{t+1} | s_t, a_t)$.

%%%%%%%%%%%%%%%%%%%%%%%%%%%%%%%%%%%%%%%%%%%%%%%%%%%%%%%%%%%%%%%%%%%%%%%%%%%%%%%%%
\section{Inverse Reinforcement Learning}\label{sec:irl}

We give a brief overview of the standard assumptions existing IRL methods make of the observation data, and mention the main approaches to inference. For a more complete review see, for example, \cite{gao2012survey}.

\subsection{Model Assumptions}

The standard IRL modeling assumption is that an agent is interacting with an MDP environment, demonstrating optimal behavior over $N$ independent episodes, thus creating paths $\Xi = (\xi_1, \dots, \xi_N)$.
Each path is a sequence of states and actions, denoted as $\xi_i = (s^i_0, a^i_0, \dots, a^i_{T_i-1}, s^i_{T_i})$, where $s_t$ and $a_t$ are the state and action at timestep $t$, and $T_i$ is the length of trajectory $i$.

An MDP $M$ is defined by the tuple $(S, A, T, R, \gamma)$, where $S$ is a set of states, $A$ is a set of actions, $T = P(s_{t+1} | s_t, a_t)$ is the transition function, $R(s)$ is the reward function, and $\gamma$ is the discount rate.
$M$ is defined in terms of some unknown parameters $\theta$.
An instance of $M$ with fixed parameters $\theta$ is denoted by $M_\theta$.

If the agent has partial observability of the environment state, the situation is defined as a POMDP $(S, A, T, R, \Omega, O, \gamma)$, where $\Omega$ is the set of possible observations and $O = P(o_t|s_t, a_t)$ is the observation function.

%\texttt{RLSIM}$(M_\theta, N)$ in Algorithm~\ref{alg:rlsim} is the generative model for simulating observation datasets $\Xi$.
%\texttt{RL}$(M_\theta)$ a generic subroutine for solving the optimal policy $\pi^*$ for $M_\theta$, and \texttt{SIM}$(M_\theta, \pi)$ a subroutine for simulating a path $\xi$ given a policy $\pi$.
%Extension to POMDP is straightforward in case the environment is only partially observable.

\subsection{Observation Assumptions}

Regarding the observations the \emph{external observer} has of the agent's behavior, four types of settings have been studied:

(1) The policy $\pi = P(a_t|s_t)$ of the agent is known
\citep{ng2000algorithms}; in other words, we know exactly how the agent will behave in any situation.

(2) Noise-free observations of the states of the environment (belief states in POMDP situations \citep{choi2011inverse}) and actions of the agent are available
\citep{ng2000algorithms,ratliff2006maximum,neu2007apprenticeship,ramachandran2007bayesian,dimitrakakis2011bayesian,rothkopf2011preference,klein2012inverse,michini2012bayesian,klein2013cascaded,tossou2013probabilistic,choi2015hierarchical,nguyen2015inverse,herman2016inverse}.
This is probably the most common formulation in the literature.
A benefit of this assumption is that it allows the likelihood to be factorized per state transition.

(3) Feature expectations of paths traveled by the agent are available
\citep{abbeel2004apprenticeship,ziebart2008maximum,boularias2011relative,bloem2014infinite}.
Feature expectations are computed from the true paths by $\hat{\mu}_E = \dfrac{1}{N}\sum^N_{i=1}\sum^{T_i}_{t=0}\gamma^t\phi(s^i_t)$, where $\phi$ is a function yielding a vector of state features.
If the reward function is linear in state features, $R(s) = \theta^T \phi(s)$, the inference problem can be formulated as a function of $\theta^T \hat{\mu}_E$.

(4) Probabilistic observations of the states of the environment are available
\citep{kitani2012activity,surana2014unsupervised}.
Here it is assumed that instead of observing the state $s_t$, the external observer only observes a distribution $u_t = P(s_t)$.
This is a natural assumption, for example, assuming measurement noise.
The general approach is to estimate the state visitation frequencies based on the observations and use them in turn to estimate the feature expectations $\hat{\mu}_E$, after which standard methods can be used.
Both feature expectations and probabilistic observations can be seen as specific summaries, or incomplete versions, of the actual paths.

\subsection{Inference Approaches}

There are two common approaches for solving the IRL problem.
MCMC can be applied for computing samples of the posterior when the unnormalized likelihood can be evaluated in closed form \citep{ramachandran2007bayesian}.
Gradient descent can be applied for giving point estimates when the gradient of the likelihood can be evaluated in closed form \citep{ziebart2008maximum}.
Also point estimation based on linear programming \citep{ng2000algorithms} and classification \citep{klein2012inverse} have been considered.

\subsection{Relationship to Imitation Learning}

The formulation of the IRL problem is close to that of imitation learning (IL), also known as apprenticeship learning \citep{abbeel2004apprenticeship}.
While in IRL we are interested in recovering the underlying parameters of the model, in IL being able to replicate the behavior of the expert is sufficient.
Thus, the goal is to recover a policy $\pi = P(a_t|s_t)$ such that the behavior generated by the policy matches that demonstrated by the expert, instead of explicitly recovering the parameters $\theta^*$ of the underlying MDP.

In general, IRL is a more complex problem than IL, as the parameter recovery problem is generally under-determined, and, depending on the formulation, may also have degenerate solutions (such as a reward function that is 0 everywhere) \citep{ng2000algorithms}.
For this reason, the approach has been to either recover the full posterior that quantifies our uncertainty \citep{ramachandran2007bayesian}, or to find point estimates that are maximally robust \citep{ratliff2006maximum}.
A solution to the IRL problem generally solves the corresponding IL problem, and might give a robust solution as the reward structure is often more generalizable compared to just a policy replicate.
For example, it is not clear how an IL policy should behave in a state that is not covered by the examples, while the parameters recovered by IRL can be used to estimate the corresponding Q-values and thus generate behavior that best follows the values of the expert.

%%%%%%%%%%%%%%%%%%%%%%%%%%%%%%%%%%%%%%%%%%%%%%%%%%%%%%%%%%%%%%%%%%%%%%%%%%%%%%%%%
\section{IRL from Summary Data}\label{sec:sdirl}

\subsection{Problem Definition}

Let $M$ be an MDP parametrized by $\theta$, where $\theta$ is any finite set of parameters of interest (not limited to the reward function parameters).
Let the true parameters be $\theta^*$ and assume an agent whose behavior agrees with an optimal policy for $M_{\theta^*}$.
We do not know $\theta^*$, but may have a prior $P(\theta)$.
Assume that the agent has taken paths $(\xi_1, \dots, \xi_N)$ but we only have observed \emph{summaries} of these paths: $\Xi_{\sigma} = (\xi_{1\sigma}, \dots, \xi_{N\sigma})$, where $\xi_{i\sigma} \sim \sigma(\xi_i)$.
$\sigma(\xi_i) = P(\xi_{i\sigma}|\xi_i)$ is a stochastic summary function that transforms a path into another type of observation, which generally contains less information than the original path (thus the name summary function).
The \emph{inverse reinforcement learning problem from summary data (IRL-SD)} problem is:\\
\textbf{Given} (1) a set of summaries $\Xi_{\sigma}$ from optimal behavior;
(2) a summary function $\sigma$;
(3) an MDP $M$ with $\theta$ unknown;
and optionally (4) a prior $P(\theta)$.\\
\textbf{Determine} $\hat{\theta}$ or the posterior $P(\theta|\Xi_\sigma)$.

In the traditional IRL setting $\theta$ would be the parameters of the reward function.
Our formulation extends the inference problem to other parameters of the MDP as well.
A similar extension in the traditional IRL setting was recently considered by \cite{herman2016inverse}.

\subsection{Motivating Example} \label{motivating_example}

To illustrate the issue with traditional IRL methods, consider the following example:
\emph{``Alice can travel from home to work using any reasonable route.
The different routes go through different kinds of scenery, and Alice has specific preferences for what kind of scenery she prefers to look at when commuting.
If we know the duration of the commute, can we say anything about Alice's preferences regarding scenery?''}

This is clearly an IRL-type problem, as the reward function of a rational agent should be estimated based on observation data.
However, all the existing methods for IRL fail to solve the problem, as no state-action trajectories or feature expectations are available.
In comparison, humans are generally able to perform inference in similar settings based on mental simulation \citep{gallese1998mirror}.
This suggests that problems such as this are regularly encountered in realistic settings and that they can be solved at least approximately in reasonable time.
%Still, this is clearly an IRL problem, as the preferences of an intelligent agent are being inferred based on observations of its behavior.

However, the above example precisely corresponds to the IRL-SD problem, with $\sigma$ extracting the duration of the path.
Thus, methods that are able to solve the IRL-SD problem will both extend the scope of problems which can be solved with IRL-type approaches and be a step towards being able to imitate human reasoning more closely.

% The example illustrates four key differences:
% First, humans do not need to observe the trajectory $\xi = (s_0, a_1, s_1, \dots, a_T, s_T)$ followed by Alice when commuting in order to make inferences relating to the commuting situation.
% Second, humans are able to make these inferences based on a minimal amount of observation data.
% Third, humans are able to come up with multiple alternative hypotheses and compare their relative plausibility.
% Fourth, humans are able to make inferences both regarding the motivations and goals of the agent (hypothesis 1), the capabilities of the agent (hypothesis 2) and the dynamics of the environment (hypothesis 3).

%There are multiple situations where we would like to estimate the parameters of a RL model, but only have access to summary-type observations. For example:

%To make the limitations that lead to IRL-SD type situations concrete, we list five specific reasons for not having access to full trajectories of behavior.

\subsection{Reasons for Summarized Observation Data}

There are multiple concrete reasons that prevent the use of full paths in modeling strategic behavior.

First, environmental and physical restrictions, such as physical occlusion or sensor saturation may prevent us from observing the full paths.

Second, coarse-grained or noisy observations are generally cheaper to acquire compared to accurate path observations.
For example, it is significantly easier to log keyboard and mouse clicks from computer users compared to eye-tracking or think-aloud observations.

Third, full path data takes up more space than summaries, which makes it more likely that only the most relevant features of the data are stored for later analysis.
Also bandwidth restrictions might prevent transmitting full path data if observations are done remotely.

Fourth, when modeling an adversary, she will likely prevent us from observing the full paths.
For example in games of incomplete information, such as poker or Starcraft, the opponent hides the details of her states and actions when possible.

Fifth, privacy guarantees result in data being released only as non-identifying summaries.
This is complementary to the previous; here the data is summarized to prevent a possible adversary from identifying specific types of information.

%%%%%%%%%%%%%%%%%%%%%%%%%%%%%%%%%%%%%%%%%%%%%%%%%%%%%%%%%%%%%%%%%%%%%%%%%%%%%%%%%
\section{Inference Methods for IRL-SD}

We first derive the observation likelihood for the IRL-SD problem.
However, as evaluating the likelihood function can be very expensive, we also propose approximations that are faster to evaluate.

\subsection{Exact Likelihood}

To derive a computable likelihood, we assume both $|S|$ and $|A|$ are finite (e.g. through discretization) and that the maximum number of actions within an observed episode is $T_{max}$.
We denote the finite set of all plausible trajectories (that have non-zero contribution to the likelihood) by $\Xi_{ap} \subseteq S^{T_{max}+1} \times A^{T_{max}}$.

The likelihood for $\theta$ given summary observations $\Xi_\sigma$ is
\begin{equation*}
  L(\theta|\Xi_\sigma)
  = \prod_{i=1}^N
    P(\xi_{i\sigma}|\theta)
  = \prod_{i=1}^N
    \sum_{\xi_i \in \Xi_{ap}}
      P(\xi_{i\sigma}|\xi_i) P(\xi_i | \theta),
\end{equation*}
where $P(\xi_{i\sigma}|\xi_i)$ is determined by the summary function $\sigma$, which is assumed to be known, and
\begin{equation*}
  P(\xi_i | \theta) = P(s^i_0) \prod_{t=0}^{T_i-1}
    \pi^*_\theta (s^i_t, a^i_t) P(s^i_{t+1} | s^i_t, a^i_t).
\end{equation*}

The main difficulty with the exact likelihood is finding the set $\Xi_{ap}$ and evaluating the sum over it.
If $\sigma$ has a known finite support, this might be used to constrain the set $\Xi_{ap}$ as paths outside the support can be immediately ruled out.

\subsection{Monte-Carlo Estimate of Likelihood}

One possibility to deal with the sum over $\Xi_{ap}$ is to use a Monte-Carlo estimate.
In this approach, paths $\Xi_{MC}$ (set of size $N_{MC} \ll |\Xi_{ap}|$) are simulated using an optimal policy $\pi^*_\theta$, so that each path is drawn with probability $P(\xi | \theta)$.
The likelihood of each individual observation can be estimated by a Monte-Carlo sum:
\begin{align*}
\hat{L}(\theta|\Xi_\sigma) =& \prod_{i=1}^N
    \dfrac{1}{N_{MC}} \sum_{\xi_n \in \Xi_{MC}}
      \dfrac{P(\xi_{i\sigma}|\xi_n) P(\xi_n | \theta)}{P(\xi_n | \theta)}\\
      =& \prod_{i=1}^N 
    \dfrac{1}{N_{MC}} \sum_{\xi_n \in \Xi_{MC}}
      P(\xi_{i\sigma}|\xi_n).
\end{align*}
As the contribution of each sample is weighted by the probability of the path, this cancels out the existing term from the product.

A benefit of this approach is that the transition probabilities $P(s_{t+1} | s_t, a_t)$ do not need to be defined any more in closed form: for generating the Monte-Carlo samples it is enough that we can draw samples.
We also need not assume that $A$ or $S$ are finite in size.

One issue with this approach is that there might not be any paths in the Monte-Carlo sample that have a non-zero observation probability for a certain observation in the dataset (that is, $P(\xi_{i\sigma}|\xi_n) = 0$ for all $n$).
This is common when $\sigma$ has a negligible support in $\Xi_{ap}$, or when the path distribution has a ``fat tail'' which is not sufficiently covered by the finite sample.
One way to alleviate this problem is to add a small constant value to the likelihood of each observation as an \emph{a-priori} estimate.
For example, $1/N_{MC}$ might be a sensible heuristic, as it vanishes with a large enough sample.

\subsection{ABC Estimate of Likelihood}

A third alternative is to avoid evaluating the likelihood function entirely, and use an approximate Bayesian computation (ABC) approach \citep{sunnaker2013approximate} instead.
ABC also uses Monte-Carlo samples for estimating the likelihood, but does it by comparing the samples directly to the observation data using a \emph{discrepancy function}, which is often chosen to be similar to the prediction error function.
Essentially this means that the Monte-Carlo sample is transformed into simulated summary observations using $\sigma$, after which the discrepancy to the observation data is computed.

The discrepancy function is denoted by
\begin{equation*}
    \delta(\Xi^{A}_\sigma, \Xi^{B}_\sigma) \rightarrow [0, \infty).
\end{equation*}
As we make no assumptions about the type of the summary observations, the choice of $\delta$ is not fixed here.
Often in the ABC literature $\delta$ is a norm between the general features of the summary datasets, or the prediction error function or its logarithm is used.

Using $\delta$ we can define a stochastic variable
\begin{equation*}
    d_\theta \sim \delta(\Xi^{sim}_\sigma, \Xi_\sigma),
\end{equation*}
where $\Xi^{sim}_\sigma = \{\sigma(\Xi_{MC,n})\}_{n=1\dots|\Xi_\sigma|}$.
The ability of $\theta$ to generate data similar to the observation data is quantified by the distribution of $d_\theta$.

The likelihood can be retrieved exactly using a $\delta$ with the property $\delta(\Xi^{A}_\sigma, \Xi^B_\sigma) = 0 \Leftrightarrow \Xi^{A}_\sigma = \Xi^B_\sigma$.
In this case the likelihood can be written as
\begin{align*}
L(\theta | \Xi_\sigma) &= P(\Xi_\sigma | \theta)
                       = P(\Xi^{sim}_\sigma = \Xi_\sigma | \theta)\\
                       &= P(d_\theta = 0 | \theta),
\end{align*}
which follows from the fact that the process for generating $\Xi^{sim}_\sigma$ is precisely our assumed generative model.

However, estimating $P(d_\theta = 0 | \theta)$ from a finite Monte-Carlo sample is challenging as most realizations lead to $d_\theta \gg 0$.
For this reason, we do an ABC approximation:
\begin{equation*}
    \tilde{L}_\varepsilon(\theta | \Xi_\sigma) = P(d_\theta \leq \varepsilon | \theta),
\end{equation*}
with an approximation threshold $\varepsilon \in [0, \infty)$.
This approximate likelihood is easier to estimate when $\varepsilon$ is similar to observed values of $d_\theta$.
The choice of $\varepsilon$ is often done adaptively.

This approach can be seen as ``IRL through imitation learning'', as we are estimating the parameter likelihood through behavior similarity.
This is an extension to matching feature expectations \citep{abbeel2004apprenticeship}, but generalized to the global features of the behavior available through $\sigma$.
A further benefit of this approach is that the observation probabilities $P(\xi_{\sigma}|\xi)$ do not need to be available in closed form, as long as we can draw samples from $\sigma$.

\subsection{Inference}

Recent work has shown the feasibility of Gaussian process (GP) \citep{rasmussen2004gaussian} surrogates for expensive likelihoods \citep{rasmussen2003gaussian}, also in the ABC setting \citep{gutmann2015bayesian}.
We use this approach as well, as the likelihoods we work with are expensive to evaluate.
The Bayesian optimization (BO) \citep{brochu2010tutorial} sampling strategy is used for concentrating the samples so that high likelihood regions are well estimated.

Algorithm~\ref{alg:inference} summarizes the estimation of the likelihood surface based on both the exact and approximate methods.
As we are performing global non-convex optimization, we make the additional assumption that the likelihood is mainly contained within a bounded region $\Theta$.
We utilize two generic subroutines:
\texttt{RL}$(M)$ is a function that given MDP $M$ finds an optimal policy $\pi^*$, and \texttt{SIM}$(M, \pi)$ is a function that given an MDP $M$ and policy $\pi$ simulates a path $\xi$ using the policy.
For a GP fit with data $D$ and hyperparameters $H$, we denote the predicted mean at $\theta$ by $G_\mu(\theta | D, H)$ and the standard deviation by $G_s(\theta | D, H)$, and
the full GP posterior is denoted by $G(\theta | D, H)$.
We denote the number of samples for estimating the surrogate by $N_{opt}$ and the BO acquisition function value at $\theta$ by $Acq(\theta | D, H)$ (the maximum of $Acq$ defines the next sample location in BO).
$\Phi(\varepsilon|\mu, \sigma)$ denotes the CDF of $N(\mu, s)$ at $\varepsilon$.
The threshold $\varepsilon$ was set to the minimum predicted value of discrepancy, as it represents the ``best that the model can do'' given the available information.

\begin{algorithm}[htb]
   \caption{Likelihood Estimation for IRL-SD}
   \label{alg:inference}
\begin{algorithmic}
  \STATE {\bfseries Input:} $M$, $\Xi_{\sigma}$, $\Theta$, $H$, $N_{opt}$, $N_{MC}$
  \STATE {\bfseries Output:} Likelihood estimate $\bar{L}(\theta)$
  \STATE $D \leftarrow \varnothing$
  \FOR{$i=1$ {\bfseries to} $N_{opt}$}
    \STATE $\theta_i \leftarrow \arg\max_\theta Acq(\theta | D, H)$
    \STATE $\pi^*_{\theta_i} \leftarrow$ \texttt{RL}$(M_{\theta_i})$
    \IF{Exact}
      \STATE $d_\theta \leftarrow \log L(\theta_i|\Xi_{\sigma})$ 
    \ELSE
      \STATE $\Xi_{MC} \leftarrow \{$\texttt{SIM}$(M_{\theta_i}, \pi^*_{\theta_i})\}_{n=1\dots N_{MC}}$
      \IF{Monte-Carlo}
        \STATE $d_\theta \leftarrow \log \hat{L}(\theta_i|\Xi_\sigma)$
      \ELSIF{ABC}
        \STATE $\Xi^{sim}_{\sigma} \leftarrow \{\sigma(\Xi_{MC,n})\}_{n=1\dots N_{MC}}$
        \STATE $d_\theta \leftarrow \delta(\Xi^{sim}_{\sigma}, \Xi_{\sigma})$
      \ENDIF
    \ENDIF
    \STATE $D \leftarrow \{D, (\theta_i, d_\theta)\}$
  \ENDFOR
  \IF{ABC}
    \STATE $\varepsilon \leftarrow \min_\theta G_\mu(\theta | D, H)$
    \STATE $\bar{L}(\theta) \leftarrow \Phi(\varepsilon | G_\mu(\theta | D, H), G_s(\theta | D, H))$
  \ELSE
    \STATE $\log \bar{L}(\theta) \leftarrow G_\mu(\theta | D, H)$
  \ENDIF
\end{algorithmic}
\end{algorithm}

For posterior inference, the log-likelihood in Algorithm~\ref{alg:inference} can be replaced with the log-posterior.
With ABC, the likelihood can be multiplied by the prior after estimation.

%%%%%%%%%%%%%%%%%%%%%%%%%%%%%%%%%%%%%%%%%%%%%%%%%%%%%%%%%%%%%%%%%%%%%%%%%%%%%%%%%
\section{Experiments}\label{sec:exp}

To study the performance of the proposed inference methods, we start with a well-known toy MDP, but change the observation assumptions to match the IRL-SD problem.
Through this example, we demonstrate that we are able to infer the parameters of the agent's reward function based only on summarized path observations.
With this MDP the approximate methods are able to recover the reward function parameters with comparable quality to the exact method, but considerably faster.

We also demonstrate that our approach scales to realistic modeling cases as well.
We show that the ABC approximation is able to infer a reasonable approximate posterior for a RL-based cognitive model from the HCI literature, based on measurements of real user behavior.
Details of the experiments are collected in Appendix~\ref{appendix}.

\subsection{Grid World}

Grid world is a well-known problem type in the IRL literature \citep{ng2000algorithms,abbeel2004apprenticeship,neu2007apprenticeship,boularias2011relative,herman2016inverse}.
In this problem, an agent is located on a cell in a discrete two-dimensional grid of $w \times w$ cells.
When the agent enters a cell, it receives a reward based on the features of the cell $\phi(s)$ and the features of the agent's reward function $\theta$, according to $R(s) = \theta^T \phi(s) + r_{step}$.

\begin{figure}[ht]
  \centering
  \includegraphics[width=0.5\columnwidth]{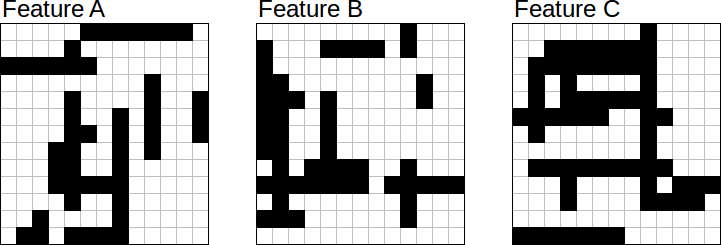}
  \caption{Visualization of a 13$\times$13 grid with three features, generated by placing 13 random walls per feature.
  Each feature is shown individually, with black squares denoting the presence of the feature.
  Each feature can be thought of as a different type of terrain (e.g. mountains, swamp, forest).
  }
\label{fig:grid}
\end{figure}

In our case, the agent is initially located on a random cell at the edge of the grid.
The cell at the center of the grid is the goal, and entering the goal gives the agent a large positive reward and ends the episode.
Each grid cell has $N_f$ binary features, which have been generated by placing $w$ walls for each feature at random on the grid (the seed value used for generating the grid is part of the MDP definition).
An example of a grid with three features is shown in Figure~\ref{fig:grid}.

The summary function is defined as $\sigma(\xi) = (s_0, |\xi|)$, yielding the initial state at the edge, and the number of steps it took to reach the goal at center (i.e. we do not know what the intermediate states or actions were).
Our problem is to infer likely values for $\theta \in [-1, 0]^{N_f}$, such that the simulated behavior with these values matches the observations, given a set of summary observations $\Xi_\sigma$ and the MDP definition.
It is also easy to verify that this corresponds to the motivating example mentioned before in Section~\ref{motivating_example}, related to Alice's scenery preferences while commuting.

\subsection{Experiment 1: Algorithm Run-time}

First, we compared the empirical run-times of the exact and approximate methods.
For the approximate methods we use a Monte-Carlo sample of size 1000.

We simulated observation sets with $N =$ 200 from grids of various sizes.
We used grids with no features ($N_f$ = 0) to avoid long paths that would make the exact method infeasible to evaluate.
We computed the first iteration step for all algorithms and recorded the elapsed wall-clock time.
The algorithms were implemented with Python and executed on an Intel Xeon X5650 2.67 GHz processor restricted to 300 MB of memory.

\begin{figure}[ht]
  \centering
  \centerline{\includegraphics[width=0.5\columnwidth]{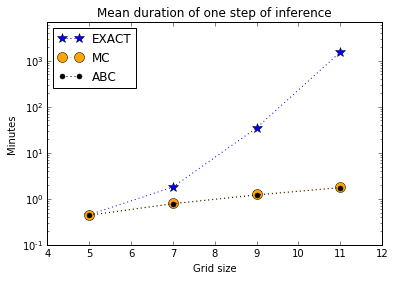}}
  \caption{Mean duration (log10 scale) of the first step of the exact and approximate methods as a function of size of the problem (N=5).
  Smaller is better.}
  \label{fig:one_step_duration}
\end{figure}

The empirical run-time of the exact algorithm grows rapidly as the size of the grid increases (Figure~\ref{fig:one_step_duration}).
This is expected, as $|\Xi_{ap}|$ grows exponentially as the length of the path grows linearly.
On the other hand, the run-times of the approximate algorithms scale comparatively much better.
ABC is equally expensive to Monte-Carlo (MC), as expected.

 % TODO: one experiment went overtime, using a conservative estimate here for that datapoint (11x11 grid)

\subsection{Experiment 2: Inference Quality}

We compared the quality of inference between the exact and approximate methods on small grids.
We also investigate the performance of the approximate methods on larger grids, where the exact method is computationally infeasible.
The experiments were performed with $N_f =$ 2 and 3.
When comparing to the exact method ($w$ being 9 and 11), we limited the length of paths in the observation dataset to be at most 12 to keep the computation time feasible (leaving on average 97~\% and 93~\% of observations, respectively).
We also use a random baseline, which is a uniform random draw from the parameter space.

We measure inference quality both by accuracy of parameter recovery, which quantifies IRL performance, and prediction accuracy, which quantifies imitation learning performance.
Accuracy of parameter recovery was measured with RMSE between likelihood mean (computed using MCMC) and ground truth.
Mean was used instead of ML as the likelihoods were sometimes broad; the mean was a more robust estimate in initial trials.

Prediction error was measured with the MAE in path length per individual starting location, measured on a separate dataset generated with the same ground truth parameters.
As discrepancy $\delta$ we used the logarithm of the prediction error computed on the observation dataset (as the errors appeared to be log-normally distributed).

\begin{figure}[htb]
  \centering
  \includegraphics[width=0.8\columnwidth]{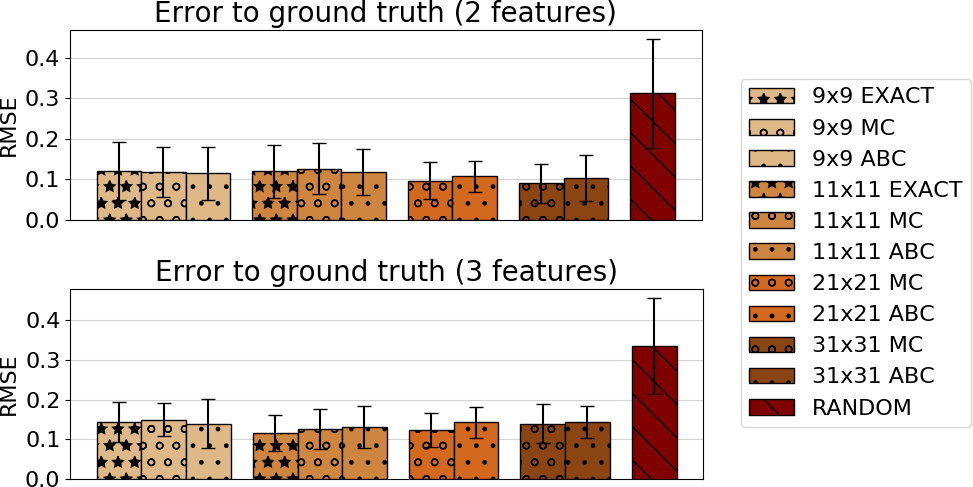}
  \caption{RMSE to ground truth (mean and standard deviation, N=30), smaller is better.}
  \label{fig:comp_gt}
\end{figure}

We observe that the approximate methods perform well compared to the exact method.
The approximate methods are able to recover the reward function parameters with comparable accuracy as the exact method, shown in Figure~\ref{fig:comp_gt}.
This demonstrates that Monte-Carlo sampling is a feasible approach for estimating the true likelihood, as is directly matching the global features of the predicted behavior with ABC.
Also, the discrepancy of the predicted behavior is relatively low with all methods, suggesting that the policies recovered by the methods are good approximations of the true policy.
There were no statistically significant differences in ground truth errors or prediction errors with any of the methods, except for the random baseline which was worse (N=30).

\begin{figure}[htb]
  \centering
  \includegraphics[width=0.8\columnwidth]{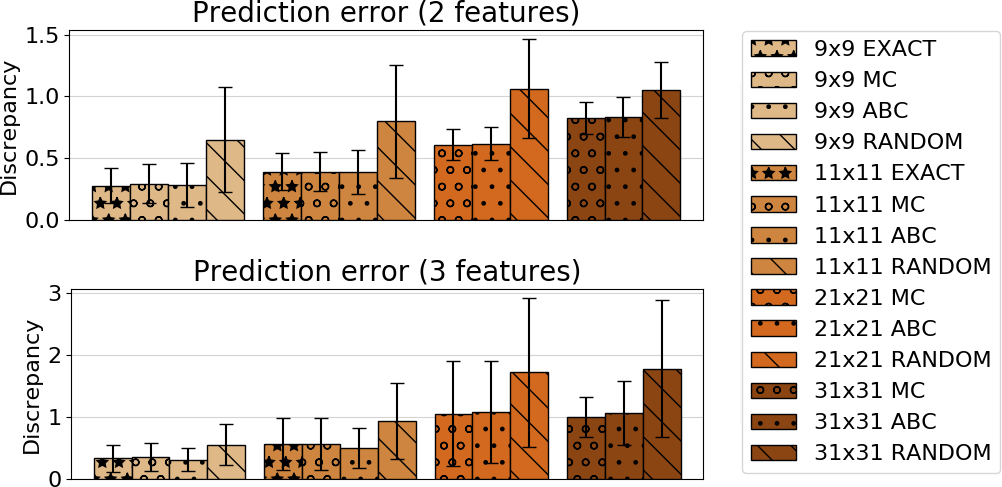}
  \caption{Prediction error on test data (mean and standard deviation, N=30), smaller is better.}
  \label{fig:comp_disc}
\end{figure}

The approximate methods are able to perform well on larger grids where the exact method is computationally infeasible.
They are able to recover the parameter values reliably (Figure~\ref{fig:comp_gt}) and the discrepancy also increases predictably with the grid size (Figure~\ref{fig:comp_disc}).

\begin{figure}[htb]
  \centering
  \includegraphics[width=0.8\columnwidth]{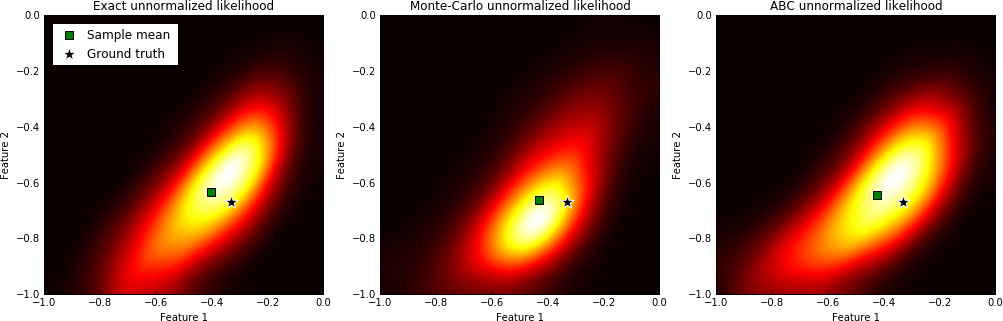}
  \caption{
  Representative example of likelihood densities estimated with different methods (2 features).
  Both Monte-Carlo and ABC are able to produce a reasonable approximation of the exact likelihood.
  Left: Exact. Center: Monte-Carlo. Right: ABC.
  The color maps are chosen so that the maxima of the functions are white and minima are black.
  The likelihood mean is marked with a square and the ground truth parameters with a star.
  More examples are found in Appendix~\ref{appendix}.}
  \label{fig:comp_logl}
\end{figure}

We also observe that the approximate likelihood densities are sensible estimates of the true likelihood, as shown in Figure~\ref{fig:comp_logl}.
In this particular example it can be seen that the ratio of the rewards is well identified, but there is still uncertainty left in the scale of the rewards.
It would not have been possible to infer this insight from just a point estimate, which demonstrates the benefit of estimating the full likelihood surface.

\subsection{Experiment 3: Modeling Computer Users}

In the final experiment we infer the full posterior of a recent RL-based cognitive model using realistic observation data.

The task is to estimate the parameters of an MDP modeling the oculomotor system of a user who is searching for a specific menu-item from a computer drop-down menu \citep{chen2015emergence,kangasraasio2016inferring}.
Although with large computer screens traditional IRL methods have been used as detailed actions can be measured with eye-tracking \citep{mohammed2015learning}, with small menus the accuracy of eye-tracking is often poor in comparison.
However, simple summary statistics, such as the time between opening a menu and clicking the target item, are simple to measure accurately, but require solving the IRL-SD problem.

Recently \cite{kangasraasio2016inferring} found MAP parameter estimates for the model using summary observations from a user study by \cite{bailly2014model}.
The summary observation included the task completion time in milliseconds (TCT, sum of the durations of all actions in an episode) and whether the target was present or absent in the menu.
We extend their analysis by showing that full posteriors can be estimated based on the same dataset and a similar model (see Appendix~\ref{appendix} for details of the model).

Although the state transition function is only defined as a computable algorithm, and the summary function $\sigma$ is a delta distribution, the ABC method is still applicable.

Getting the average TCT predicted correctly is the primary goal of the model, and getting the variation correct as well is the secondary goal.
For this reason, the discrepancy function $\delta$ was chosen to be logarithm of the squared differences in TCT means plus absolute differences in standard deviations summed from both menu conditions.

We infer the posteriors of three parameters of the MDP:
 (1) the duration of eye fixations $f_{dur}$ (units of 100 ms);
 (2) the duration of moving the mouse to select an item $d_{sel}$ (units of 1 s); and
 (3) the probability of recalling the full menu layout from memory $p_{rec}$.

The reward function is such that the agent receives a penalty equal to the milliseconds spent on performing the action.
The duration of an action is the sum of saccade duration (based on the distance between two consecutive fixation locations), $f_{dur}$ and $d_{sel}$.
From this perspective, $f_{dur}$ and $d_{sel}$ can also be seen as parameters of the reward function.
Finding the correct item leads to a reward 10k, as does quitting when there is no target item in the menu.
Quitting when there is target present results in a penalty -10k.

\begin{figure}[ht]
  \centering
  \centerline{\includegraphics[width=0.8\columnwidth]{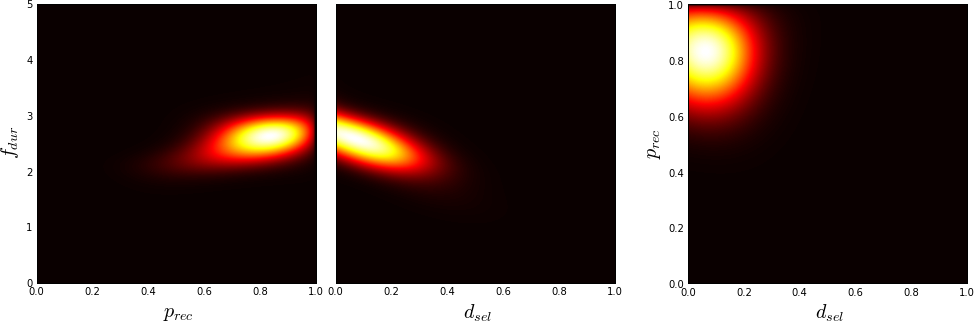}}
  \caption{The approximate posterior inferred with ABC demonstrates that the parameters can be identified and that the remaining uncertainty is well characterized.
  Left: fixation duration $f_{dur}$ and menu recall probability $p_{rec}$.
  Center: fixation duration $f_{dur}$ and selection delay $d_{sel}$.
  Right: menu recall probability $p_{rec}$ and selection delay $d_{sel}$.
  The color map is chosen so that the maximum of the posterior is white and minimum is black.
  }
  \label{fig:menu_post}
\end{figure}

The posterior is visualized in Figure~\ref{fig:menu_post} using 2D slices at the MAP location ($d_{sel} =$ 0.05, $p_{rec} =$ 0.80, $f_{dur} =$ 2.6).
We observe that \emph{a posteriori} there is a correlation between $f_{dur}$ and $p_{rec}$, and similarly for $f_{dur}$ and $d_{sel}$.
Both of these are understandable, as increasing $f_{dur}$ would increase the predicted TCT, as would decreasing $p_{rec}$ or increasing $d_{sel}$.
The posterior of $f_{dur}$ is centered around 260~ms, but there is still uncertainly left in $d_{sel}$ and $p_{rec}$.
The uncertainty in $d_{sel}$ is explained by the difficulty of pointing precisely to the target item with the cursor, which causes variation in its duration.
The uncertainty in $p_{rec}$ is explained by the fact that the menus encountered early on in the experiments were completely new to the subjects, but as the experiment progressed the subjects were more and more likely to recall the menus.
We also observe that there is no significant posterior correlation between $p_{rec}$ and $d_{sel}$.
This indicates that although they both affect the TCT, the effects they have are orthogonal; increasing the probability of recalling a menu can not be fully compensated just by increasing the selection duration.

The simulated data at the MAP location was able to reproduce the general features of the observation data.
A comparison of key features is shown in Table~1. % for some reason refs don't work for tables?

\begin{table}[htb]
    \centering
    \begin{tabular}{ccc}
        \textbf{Feature} & \textbf{MAP} & \textbf{Observation data} \\ \hline
        TCT (abs)      & 430 ms & 470 ms \\
        TCT (pre)      & 980 ms & 970 ms \\
        Saccades (abs) & 1.4    & 1.9 \\
        Saccades (pre) & 3.1    & 2.2 \\
    \end{tabular}
    \caption{
    Comparison of menu model prediction means (MAP estimate) and observation data means.
    The condition when the target is absent from the menu is denoted by (abs), and the condition when the target is present by (pre).
    }
\end{table}

%%%%%%%%%%%%%%%%%%%%%%%%%%%%%%%%%%%%%%%%%%%%%%%%%%%%%%%%%%%%%%%%%%%%%%%%%%%%%%%%%
\section{Discussion}

The experiments demonstrate that the proposed approximate methods are applicable for inferring RL-based models based on aggregate observation data, when it is acceptable that the inference takes some time.
For example, many off-line scientific modeling scenarios fall into this setting.
However, there are still multiple complementary options for improving the speed and scalability of the proposed methods from here on.
One option to scale up to higher-dimensional parameter spaces is to find a lower-dimensional subspace where the most interesting variation takes place \citep{wang2016bayesian}.
One option to increase the speed of finding solutions to RL problems is to use RL transfer learning, as it is generally faster to find a good policy based on an existing policy from a nearby location \citep{ramachandran2007bayesian}, compared to learning it from scratch.

An interesting feature of both of the proposed approximations is that they do not explicitly depend on the path likelihood.
With the MC approximation, this is due to a term cancellation, and with the ABC approximation this is due to the likelihood-free modeling approach.
This means that the limitations to performance are different than usually;
instead of being limited by the ability to evaluate the path likelihood function, the methods are limited by the ability to generate reasonable behavior with certain parameter values.
Although generating samples from the model is often a less efficient inference method compared to evaluating the likelihood function directly, the situation is different when one does not have the luxury of choosing the observation data to precisely match the model assumptions, such that the likelihood would have a convenient form.
%Although this change is generally inefficient when the likelihood factorizes conveniently per state transition, the situation is different when one does not have the luxury of choosing the observation data to precisely match the model assumptions.
Furthermore, the fact that the generative model is now ``decoupled'' from the inference method might open up new avenues of research in modeling strategic behavior, as this decoupling enables greater flexibility in the design of the generative model, instead of being limited strictly to the MDP assumptions.

With full path observations the summary function $\sigma$ becomes the identity and the exact likelihood becomes the same as in most traditional IRL methods.
Thus the proposed exact method should in principle yield a similar posterior as existing Bayesian IRL methods \citep[e.g.][]{ramachandran2007bayesian}.
The two proposed approximations have been designed specifically to the situation where the observations are available only in summarized form.
In principle, the MC approximation works best when the image of the function $\sigma$ is not ``significantly larger'' compared to the size of the Monte-Carlo sample, as this keeps the estimation noise due to the finite sample relatively small.
With full path observations, the possibility of sampling precisely similar paths as in the observation data might be arbitrarily small, which causes practical problems with this method.
The ABC approximation can be used with full path observations as long as the discrepancy function $\delta$ and threshold $\varepsilon$ are reasonable.
However, due to the likelihood-free approach, the ABC approximation will likely be slower than more specialized methods when the full paths are available and the likelihood gradient is computable.

The need to have some knowledge of the summarizing function $\sigma$ is, in general, an unavoidable requirement for performing inference.
In this work it was assumed that $\sigma$ was known in advance.
%However, in general it should be possible to infer $\sigma$ if sufficient training data is available.
If $\sigma$ is unknown, it might be estimated from data if full path observations are available for some data.

Also, it is clear that the amount of information available of the model parameter values depends on $\sigma$.
Thus, not all possible $\sigma$ lead to a feasible setting for inference.
As it is challenging to define requirements for $\sigma$ without considering the specific application, evaluating the feasibility of inference needs to be made based on expert knowledge or empirical experiments.
However, a key benefit of the proposed Bayesian approach is that the full posterior allows the remaining uncertainty to be directly estimated.

The need to choose the discrepancy function $\delta$ and threshold $\varepsilon$ is unavoidable in ABC; a recent summary of different methods is provided by \cite{lintusaari2017fundamentals}.
The most promising choices are to either use domain knowledge, which is naturally task-specific, or more generally to learn from data a classifier which can be used to form the discrepancy function \citep{gutmann2014statistical}.

%%%%%%%%%%%%%%%%%%%%%%%%%%%%%%%%%%%%%%%%%%%%%%%%%%%%%%%%%%%%%%%%%%%%%%%%%%%%%%%%%
\section{Summary}

In this paper we defined the IRL-SD problem, where the task is to do inverse reinforcement learning based on summarized observations of the agent's behavior.
We proposed exact and approximate methods for inference.
The Monte-Carlo approximation can be used when the summary function $\sigma$ is available as a probability distribution with a non-negligible support, and the ABC even when $\sigma$ can only be evaluated.
We demonstrated that all proposed methods are able to produce feasible results, but the exact method is computationally expensive.
However, the approximate methods can be used even for full posterior inference with realistic MDPs and real observation data.
The methods presented are feasible baselines for more specialized inference algorithms that may take advantage of further assumptions, and are state-of-the-art in situations that are currently out-of-reach for existing more specific methods.

Overall, regarding partial observability in IRL, there have been two cases for which methods exist:
\begin{itemize}
    \item If the agent has partial observability of the environment state, a POMDP model can be used \citep{choi2011inverse}.
    \item If the external observer has partial observability of the environment state, traditional IRL methods can be extended \citep{kitani2012activity}.
\end{itemize}
This work extends this list by a third item:
\begin{itemize}
    \item If the external observer has partial observability of the complete path, then the presented methods for IRL-SD can be applied.
\end{itemize}

\begin{acknowledgements}
This work has been supported by the Academy of Finland (Finnish Centre of Excellence in Computational Inference Research COIN, and grants 294238, 292334).
Computational resources were provided by the Aalto Science IT project.
\end{acknowledgements}

\bibliographystyle{spbasic}      % basic style, author-year citations
%\bibliography{references.bib}

\appendix

\section{Supplementary Material}
\label{appendix}

\subsection{Experiments 1 and 2}

\noindent \textbf{Model Details}

\noindent The walls that define the grid cell features are generated as follows:
(1) choose one grid cell at random;
(2) choose vertical or horizontal direction at random;
(3) choose another cell at random along the chosen direction;
(4) set the feature values to 1 for all cells between (and including) these two cells, except if the cell is the goal cell.

The agent has four actions that allow it to move to neighboring cells.
Each action fails with probability $p_{slip}$, resulting in the agent moving to a random neighboring cell.
Attempting to move outside the grid returns the agent to the current cell.
We used $p_{slip} =$ 0.05 and $r_{step} =$ -0.05.
$T_{max}$ was set to 10$w$ steps.
Optimal policy was estimated with Q-learning over 2000$w$ episodes in batches of 500; parameters were step size 0.2, learning rate 0.5, $\gamma$ 0.99, exploration rate 0.2.

\phantom{.}

\noindent \textbf{Inference Details}

\noindent For the Monte-Carlo estimate of the likelihood, we added $1/N_{MC}$ to the estimated likelihood of each observation, to prevent the likelihood being 0 when when $P(\xi_{i\sigma}|\xi_n) = 0$ for all $n$ in the sample set.
This can be thought of as a ``prior'' for the observation likelihood.

\begin{figure}[htb]
  \centering
  \includegraphics[width=0.7\columnwidth]{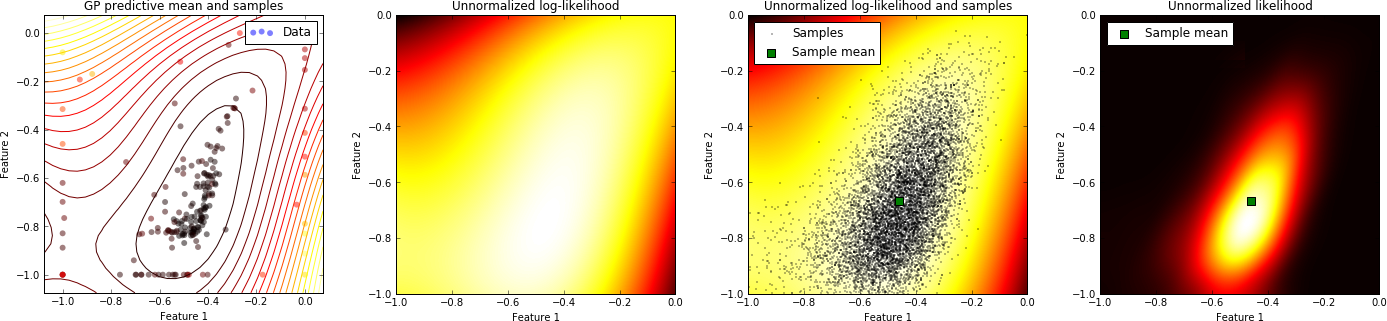}
  \caption{
  Left: GP predictive mean and BO samples.
  Center left: Unnormalized log-likelihood surface mean.
  Center right: Samples from log-likelihood and sample mean on top of unnormalized log-likelihood surface mean.
  Right: Estimated likelihood density and sample mean.}
  \label{fig:logl_construction}
\end{figure}

We estimated the mean of the likelihood using MCMC samples (Metropolis-Hastings sampling).
We drew 10000 samples with a burnout of 1000 and thinning 5, starting from the center of the boundaries.
The proposal distribution was a symmetric Gaussian with standard deviation 0.1.
For the exact and Monte-Carlo estimates, as we are sampling from a distribution over the log-likelihood surfaces (the GP posterior), for computing the acceptance ratio, for each sample, we drew one realization from the GP posterior and computed the likelihood ratio on that realization.
We found that this way of taking the uncertainty into account was superior to just using the predictive mean.
The samples from the predictive mean often resulted in an unrealistically narrow likelihood as they did not take into account the remaining uncertainty in the log-likelihood surface.
For the ABC approach, we used standard MCMC as the uncertainty in the GP surface is taken into account by the ABC likelihood.
For estimating the likelihood density from the samples, we used a Gaussian kernel density estimate with the bandwidth decided by Scott's rule.
Illustration of the process is shown in Figure~\ref{fig:logl_construction}.

%\subsubsection{Inference Details}

%As the number of possible paths given an observation is finite, the likelihood function can be evaluated in principle.
%Given a start state and path length, the paths form a tree with branching factor at most 16 and depth equal to the path length (there are at most 4 optimal actions at each state and at most 4 possible succeeding states).
%The tree can be pruned by requiring that the absorbing goal state appears exactly once in the path, as the final state.
%Also, as in this case we use a deterministic policy, we can prune out all actions for each state that disagree with the policy.

\phantom{.}

\noindent \textbf{Experiment 2 Details}

\begin{figure}[htb]
  \centering
  \includegraphics[width=0.5\columnwidth]{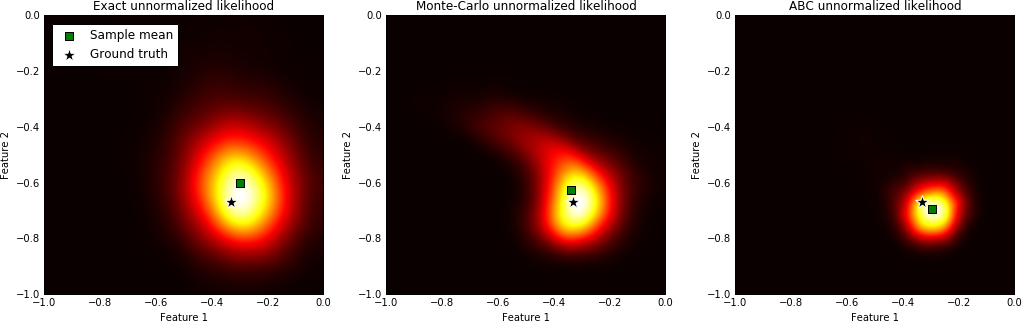}
  \caption{
  An example of a narrow likelihood.
  In this case it is possible to identify the values of both parameters with good accuracy based on the observation data.}
  \label{fig:comp_logl_1}
\end{figure}

\begin{figure}[htb]
  \centering
  \includegraphics[width=0.5\columnwidth]{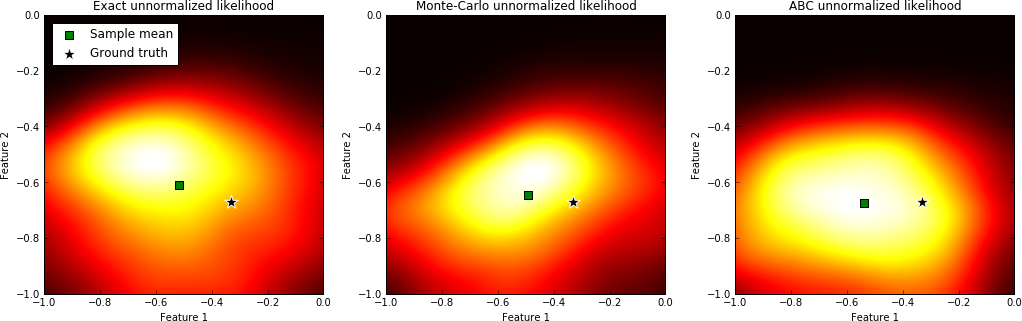}
  \caption{
  An example of a wide likelihood.
  In this case the parameter values are not well identified based on the observation data.}
  \label{fig:comp_logl_2}
\end{figure}

\noindent We used evenly spread out constant values for the parameter ground truth $\theta^*$.
For $N_f =$ 2 we used [-0.33, -0.67] and for $N_f =$ 3 we used [-0.25, -0.5, -0.75].
This was done to remove the noise in the results caused by variation in the ground truth and to promote the identifiability of the parameters.
With 2 features, the number of BO samples was 200, computed in batches of 10. With 3 features, the number of samples was increased to 600.
$\Theta$ was [-1, 0]$^{N_f}$.
The same sets of $\Xi_\sigma$ was used for all algorithms when comparing performance.
Further examples of different types of characteristic likelihood surfaces are shown in Figures~\ref{fig:comp_logl_1}~and~\ref{fig:comp_logl_2}.

% \begin{figure}[htb]
%   \centering
%   \includegraphics[width=0.7\columnwidth]{new_comp_logl_one_param.png}
%   \caption{
%   An example of an oblong likelihood.
%   In this case one of the parameter values is well identified, while the other one is not.}
%   \label{fig:comp_logl_3}
% \end{figure}

\subsection{Experiment 3}

\noindent \textbf{Model Definition}

\noindent The MDP environment contains a drop-down menu that consists of 8 items, which can be in multiple states: the semantic relevance of the item is either unobserved, low, medium or high, or the item is the target item.
The length of the item is either unobserved, correct or incorrect.
These constitute the state that the agent observes.
Initially all items are unobserved, and the agent can observe them by fixating at any item.
The agent always observes the semantic relevance of the fixated item.
The semantic relevance of a neighboring item is observed with probability $p_{sem}$.
With probability 0.95 the agent will observe the length of the current item and with probability 0.89 the length of a neighboring item.
Also, with probability $p_{rec}$ the agent will recall the full menu layout after the first fixation (that is, observe the features of all the items).
The duration of a fixation is $f_{dur}$ (in units of 100 ms). 
When the agent fixates on the target item, the agent will select it, which takes additional $d_{sel}$ seconds.
The agent can also quit, which is instantaneous.
The probability for a target item to appear in a menu was 0.9 and the maximum number of actions in a session was 100.

We note that the model is actually a POMDP encoded as an MDP, which is why the transition function is difficult to define as an explicit probability distribution.
The implementation generates a full menu layout at the beginning of each episode, and the state transitions during the episode are based on this menu realization.

\phantom{.}

\noindent \textbf{Inference}

\noindent We use a similar prior as \cite{kangasraasio2016inferring}: for $f_{dur}$ we use truncated Normal with $\mu$ 3, $\sigma$ 1; for $d_{sel}$ truncated Normal with $\mu$ 0.3, $\sigma$ 0.3, for $p_{rec}$ Beta with $\alpha$ 3, $\beta$ 1.35.
$\Theta$ was $f_{dur} \in [0, 5]$, $d_{sel} \in [0, 1]$, $p_{rec} \in [0, 1]$.
The optimal policy is estimated with Q-learning over 5M sessions in batches of 10k sessions.
Step size was 0.05, learning rate 0.3, $\gamma$ 0.98 and exploration rate 0.1.
The training was done on a fixed set of 20k menus and the predictions were done on a separate set of 10k menus.
We computed 1000 BO samples in batches of 50 for estimating the posterior.

%\subsubsection{Inference Quality}

\subsection{Bayesian Optimization Details}

A radial basis function kernel was used for the GP.
The initial lengthscale of the kernel was set to 10~\% of the bound width, the variance to roughly 50~\% of the maximum difference between the minimum and maximum sample values observed in initial tests and the noise variance to 0.1.
After each batch, the parameters were optimized from the initial values.
For computing the batch sample locations we used a LCB acquisition rule combined with local penalization \citep{gonzalez2016batch}.

% \subsection{Extension to POMDP}

% If the situation is POMDP, the exact likelihood becomes even more infeasible to evaluate, as $P(\xi_i|\theta)$ is more complex and as the belief-states are continuous.

% However, both approximations should remain feasible, as long as there exists a method for estimating $\pi^*_\theta$ (e.g. a POMDP-solver).
% The Monte-Carlo sample can be constructed using the estimated policy and the methods applied similarly as in the MDP case.

% Extending to the POMDP case will likely increase the computational cost of inference, as it is generally more difficult to find good estimates for the optimal policy.

\end{document}